# Comprehensive Overview of Artificial Intelligence Applications in Modern Industries

Yijie Weng, Jianhao Wu, Tara Kelly, William Johnson

jaweng333@gmail.com, johnwu2417@gmail.com, tarakelly963@gmail.com, william.johnson.22@gmail.com

**Abstract**

Artificial Intelligence (AI) is fundamentally reshaping various industries by enhancing decision-making processes, optimizing operations, and unlocking new opportunities for innovation. This paper explores the applications of AI across four key sectors: healthcare, finance, manufacturing, and retail. Each section delves into the specific challenges faced by these industries, the AI technologies employed to address them, and the measurable impact on business outcomes and societal welfare. We also discuss the implications of AI integration, including ethical considerations, the future trajectory of AI development, and its potential to drive economic growth while posing challenges that need to be managed responsibly.

## 1. Introduction

Artificial Intelligence (AI) has emerged as one of the most transformative technologies of the 21st century. Defined as the capability of a machine to mimic intelligent human behavior, AI encompasses a broad range of technologies, including machine learning, natural language processing, computer vision, and robotics. Its applications are far-reaching, impacting diverse fields such as healthcare, finance, manufacturing, retail, education, and more.

The proliferation of data, advancements in computational power, and the development of sophisticated algorithms have accelerated the adoption of AI across industries. Businesses are leveraging AI not only to automate repetitive tasks but also to gain insights from data, improve customer experiences, and innovate in product and service offerings.

In this paper, we provide an in-depth analysis of AI applications in four key industries: healthcare, finance, manufacturing, and retail. For each sector, we will examine the types of AI technologies being used, the problems they aim to solve, the benefits they bring, and the challenges associated with their deployment. We will also explore future trends and the broader implications of AI adoption.

## 2. Applications of AI in Healthcare

Healthcare is a field characterized by complexity, variability, and the critical need for precision. The integration of AI in healthcare has the potential to address some of the most pressing challenges, such as improving diagnostic accuracy, personalizing treatment, and optimizing healthcare delivery. This section explores how AI is being used to revolutionize medical diagnostics, drug discovery, personalized medicine, and operational efficiencies.

## 2.1 Medical Diagnostics and Imaging

Medical diagnostics, especially those relying on imaging, is one of the areas where AI has shown the most promise. Traditional diagnostic processes can be time-consuming and subject to human error, particularly in complex cases. AI algorithms, especially those based on deep learning, have demonstrated the capability to analyze medical images with remarkable accuracy, sometimes surpassing human experts.

Deep learning models, such as convolutional neural networks (CNNs), are trained on vast datasets of labeled medical images. These models can learn to recognize subtle patterns and anomalies that may not be apparent to human clinicians. For example, AI systems have been developed to detect early signs of diseases such as cancer, tuberculosis, and neurological disorders from radiographic images, MRI scans, and CT scans. In oncology, AI has been used to identify malignant tumors at an earlier stage, when treatment options are more effective.

The application of AI in medical imaging extends beyond disease detection. It is also being used for image enhancement and reconstruction, enabling more detailed and clearer images from lower radiation doses, thus improving patient safety. Companies like Enlitic, Arterys, and PathAI are at the forefront of developing AI-driven diagnostic tools that are being integrated into clinical workflows worldwide.

Despite its potential, the deployment of AI in diagnostics is not without challenges. One of the primary concerns is the need for large, diverse, and high-quality datasets to train AI models. Ensuring the generalizability of these models across different patient populations and imaging equipment is crucial. Furthermore, regulatory approval processes for AI-based medical devices need to balance innovation with patient safety.

## 2.2 Drug Discovery and Development

The traditional drug discovery process is often described as a costly and time-consuming endeavor, with a high rate of failure. On average, it takes over a decade and billions of dollars to bring a new drug from the laboratory to the market. AI is poised to revolutionize this process by enabling more efficient identification of drug candidates and predicting their success in clinical trials.

Machine learning models can analyze vast amounts of chemical and biological data to identify potential drug candidates. For instance, deep learning models can predict the biological activity of chemical compounds, helping researchers focus on the most promising candidates. Natural language processing (NLP) techniques can also be used to mine scientific literature and patents to identify novel drug targets and potential drug-repurposing opportunities.

One of the most significant breakthroughs in AI-driven drug discovery is the use of generative models to design new molecules. These models, which include variational autoencoders (VAEs) and generative adversarial networks (GANs), can generate novel chemical structures with desired properties. For example, researchers have used these models to design new antibiotics and antiviral compounds, potentially accelerating the development of treatments for diseases that currently lack effective therapies.

AI is also transforming the clinical trial phase of drug development. Predictive models can help identify patient populations that are most likely to benefit from a new drug, thereby reducing the size and duration of clinical trials. AI can also optimize trial design by predicting potential side effects and identifying biomarkers that can be used to monitor treatment efficacy.

However, the integration of AI in drug discovery and development faces several challenges. The pharmaceutical industry is heavily regulated, and the use of AI models requires rigorous validation to ensure their reliability and safety. Additionally, collaboration between AI researchers and domain experts is essential to ensure that AI models are used effectively and ethically in the drug discovery process.

## 2.3 Personalized Medicine and Genomics

Personalized medicine represents a paradigm shift in healthcare, moving away from the traditional "one-size-fits-all" approach to treatment towards more individualized care. AI is a key enabler of this shift, particularly through its application in genomics.

The human genome contains over three billion base pairs, and understanding how variations in this genetic code contribute to health and disease is a complex task. AI is helping to decode this complexity by analyzing large-scale genomic data to identify genetic variants associated with specific conditions. Machine learning models can identify patterns in genomic data that are linked to diseases such as cancer, diabetes, and cardiovascular disorders.

AI is also being used to predict individual responses to different treatments based on genetic and clinical data. This is particularly important in oncology, where the genetic profile of a tumor can influence its response to chemotherapy or targeted therapies. AI models can analyze genomic data alongside clinical records to recommend personalized treatment plans, potentially improving outcomes and reducing adverse reactions.

The integration of AI in genomics is also advancing our understanding of rare genetic disorders. Traditional methods of identifying disease-causing mutations can be labor-intensive and time-consuming. AI models can analyze whole-genome sequencing data to identify rare variants that may be responsible for these conditions, enabling earlier diagnosis and intervention.

Despite the promise of AI in personalized medicine, several challenges remain. Access to high-quality genomic data is a significant barrier, as is the need for robust methods to integrate genetic, clinical, and environmental data. Ethical considerations, such as patient privacy and consent, are also critical in the use of AI for personalized medicine.

## 3. Applications of AI in Finance

The finance industry, characterized by its data-intensive and risk-sensitive nature, has been an early adopter of AI technologies. From algorithmic trading and fraud detection to risk management and customer service, AI is transforming financial services in profound ways. This section explores how AI is being applied in finance, its impact on efficiency and decision-making, and the challenges associated with its use.

### 3.1 Algorithmic Trading and Market Analysis

Algorithmic trading, also known as algo-trading, refers to the use of computer programs to execute trades based on pre-defined criteria at speeds and frequencies that are beyond human capabilities. AI has significantly enhanced the capabilities of algorithmic trading by enabling more sophisticated strategies based on real-time data analysis.

Machine learning models, particularly those using reinforcement learning, can adapt to changing market conditions by continuously learning from historical data. These models can identify complex patterns and correlations that are not apparent through traditional statistical methods. For example, AI systems can analyze price movements, trading volumes, and even news sentiment to predict short-term market trends and execute trades accordingly.

High-frequency trading (HFT) is a subset of algorithmic trading that relies on AI to make split-second trading decisions. HFT firms use AI models to analyze market data in real-time and execute large volumes of trades in milliseconds. This speed advantage allows them to capitalize on minute price discrepancies, generating significant profits. However, the use of AI in HFT has raised concerns about market stability and fairness, as these systems can amplify market volatility during periods of uncertainty.

AI is also being used for market analysis and portfolio management. Robo-advisors, powered by AI algorithms, provide personalized investment advice based on individual risk preferences and financial goals. These systems use machine learning to optimize portfolios and rebalance assets in response to market changes. AI-driven hedge funds, such as those operated by Renaissance Technologies and Two Sigma, use similar techniques to develop complex trading strategies that have consistently outperformed traditional methods.

Despite the advantages of AI in trading, there are challenges and risks associated with its use. The "black box" nature of many AI models makes it difficult to understand how they arrive at specific decisions, which can be problematic in a regulatory context. Additionally, the reliance on historical data means that AI models may not always perform well in unprecedented market conditions, such as those caused by a financial crisis or geopolitical event.

### 3.2 Fraud Detection and Prevention

Fraud detection is a critical area where AI has proven to be highly effective. Financial fraud, including credit card fraud, identity theft, and money laundering, is a persistent and evolving threat. Traditional rule-based systems are often inadequate for detecting sophisticated fraud schemes, as they rely on predefined patterns that fraudsters can learn to bypass.

AI models, particularly those based on machine learning, can analyze vast amounts of transaction data in real-time to detect anomalies that may indicate fraudulent activity. These models can learn from both historical fraud cases and legitimate transactions to identify patterns that are indicative of fraud. For example, AI can detect unusual spending patterns, such as a sudden surge in high-value transactions from a previously dormant account, or transactions originating from multiple locations within a short period.

Natural language processing (NLP) is also being used to analyze unstructured data, such as email communications and social media posts, to detect potential fraud. AI models can flag suspicious activities, such as phishing attempts or fraudulent account takeovers, before they result in financial losses.

In the realm of anti-money laundering (AML), AI is helping financial institutions comply with regulatory requirements by automating the monitoring and reporting of suspicious transactions. Machine learning models can analyze transaction flows and customer behavior to identify potential money laundering activities, reducing the burden on human analysts and improving the effectiveness of AML programs.

However, the use of AI in fraud detection presents several challenges. False positives, where legitimate transactions are flagged as fraudulent, can be a significant issue, leading to customer dissatisfaction and increased operational costs. Additionally, as AI systems become more sophisticated, so do the tactics used by fraudsters to evade detection. Continuous model updates and the integration of new data sources are essential to maintaining the effectiveness of AI-based fraud detection systems.

### 3.3 Risk Management and Credit Scoring

Risk management is a cornerstone of the financial industry, encompassing everything from credit risk and market risk to operational risk and liquidity risk. AI is being used to enhance risk management processes by providing more accurate risk assessments and enabling proactive measures to mitigate potential losses.

In credit risk assessment, AI models can analyze a wide range of data sources, including credit history, employment records, and social media activity, to evaluate the creditworthiness of individuals and businesses. Traditional credit scoring models, such as FICO, primarily rely on historical credit data. In contrast, AI models can incorporate alternative data sources to provide a more holistic view of an applicant's financial stability. This approach has the potential to expand access to credit for individuals who may not have a traditional credit history but are nonetheless creditworthy.

AI is also being used to monitor market risk by analyzing real-time data from financial markets and identifying potential risk factors. Predictive models can simulate the impact of various scenarios, such as interest rate changes or geopolitical events, on investment portfolios. This enables financial institutions to take preemptive actions, such as rebalancing portfolios or hedging against potential losses.

In operational risk management, AI can help identify and mitigate risks associated with internal processes and systems. For example, machine learning models can analyze historical incident reports and operational data to identify patterns that may indicate potential risks, such as system failures or security breaches.

While AI offers significant advantages in risk management, there are challenges to its adoption. The interpretability of AI models is a major concern, as financial institutions must be able to explain their risk assessments to regulators and stakeholders. Additionally, the use of AI in risk management requires robust data governance frameworks to ensure the quality and integrity of the data being used.

### 4. Applications of AI in Manufacturing

Manufacturing has undergone significant transformations with the advent of AI, particularly in areas such as predictive maintenance, quality control, supply chain optimization, and the development of smart factories. This section explores how AI is being used to enhance efficiency, reduce costs, and improve product quality in the manufacturing sector.

### 4.1 Predictive Maintenance and Equipment Monitoring

Unplanned equipment downtime can have a significant impact on manufacturing operations, leading to lost production time, increased maintenance costs, and delayed order fulfillment. Traditional maintenance strategies, such as reactive maintenance and preventive maintenance, are often inadequate for minimizing downtime and optimizing maintenance schedules.

AI-powered predictive maintenance offers a solution by enabling manufacturers to anticipate equipment failures before they occur. Machine learning models can analyze data from sensors embedded in machinery to detect patterns that indicate potential failures. These models can monitor parameters such as temperature, vibration, and pressure in real-time, providing early warnings of equipment anomalies.

By predicting when and where equipment is likely to fail, manufacturers can schedule maintenance activities more effectively, reducing unplanned downtime and extending the lifespan of machinery. Predictive maintenance can also optimize the use of maintenance resources by ensuring that repairs are carried out only when necessary, rather than based on fixed schedules.

The benefits of AI in predictive maintenance extend beyond individual machines to entire production lines and facilities. For example, General Electric (GE) has developed AI-powered systems that monitor the performance of industrial equipment in real-time, enabling proactive maintenance and reducing operational disruptions.

However, implementing predictive maintenance requires significant investments in data collection and infrastructure. Manufacturers must deploy sensors and connectivity solutions to collect data from equipment, and they must have the computational resources to process and analyze this data in real-time. Additionally, integrating AI models into existing maintenance workflows can be challenging, as it requires collaboration between data scientists, engineers, and maintenance personnel.

**4.2 Quality Control and Defect Detection**

Maintaining high product quality is a top priority for manufacturers, as defects can lead to customer dissatisfaction, product recalls, and financial losses. Traditional quality control methods, such as manual inspections and statistical process control, can be time-consuming and prone to errors. AI is revolutionizing quality control by enabling automated and accurate defect detection in real-time.

Computer vision, a subfield of AI, is particularly useful for quality control in manufacturing. Machine learning models can be trained to recognize visual defects in products by analyzing images captured by cameras on production lines. These models can detect a wide range of defects, including surface blemishes, dimensional inaccuracies, and assembly errors.

AI-powered quality control systems can operate continuously, inspecting products at a speed and consistency that far exceed human capabilities. This not only improves the accuracy of defect detection but also enables manufacturers to identify quality issues earlier in the production process, reducing scrap rates and rework costs.

Beyond visual inspection, AI can be used to monitor and control other aspects of product quality, such as material properties and process parameters. For example, machine learning models can analyze data from sensors embedded in production equipment to ensure that temperature, pressure, and other variables remain within acceptable ranges. This enables manufacturers to maintain consistent product quality and comply with regulatory standards.

Despite the advantages of AI in quality control, several challenges must be addressed. Training AI models for defect detection requires large datasets of labeled images, which can be difficult to obtain in industries where defects are rare. Additionally, integrating AI systems into existing production lines may require significant modifications to equipment and processes.

**4.3 Supply Chain Optimization and Inventory Management**

Efficient supply chain management is essential for manufacturing success, as it ensures the timely delivery of raw materials and the distribution of finished products to customers. AI is being used to optimize various aspects of the supply chain, from demand forecasting and inventory management to logistics and supplier management.

Machine learning models can analyze historical sales data, market trends, and external factors such as weather and geopolitical events to forecast demand for different products. This enables manufacturers to optimize inventory levels, reducing both excess stock and stockouts. For example, AI-powered demand forecasting has helped companies like Siemens and Bosch improve their production planning and reduce inventory holding costs.

AI is also being used to optimize logistics and transportation in the supply chain. Predictive models can analyze traffic patterns, weather conditions, and shipping data to determine the most efficient routes and delivery schedules. This reduces transportation costs and improves delivery reliability.

In supplier management, AI can analyze supplier performance data to identify potential risks, such as late deliveries or quality issues. This enables manufacturers to take proactive measures, such as diversifying their supplier base or renegotiating contracts, to mitigate these risks.

While AI offers significant benefits in supply chain optimization, its adoption requires a high level of data integration and collaboration across the supply chain network. Manufacturers must work closely with suppliers, distributors, and logistics providers to share data and implement AI-driven solutions effectively. Additionally, the use of AI in supply chain management raises concerns about data privacy and security, as sensitive business information is shared across multiple stakeholders.

## 5. Applications of AI in Retail

The retail industry is undergoing a digital transformation, driven in part by the adoption of AI technologies. From personalized recommendations and dynamic pricing to inventory management and customer service, AI is helping retailers improve operational efficiency, enhance the customer experience, and increase sales. This section explores the various ways in which AI is being applied in the retail sector.

### 5.1 Personalized Recommendations and Customer Engagement

Personalization is a key differentiator in the competitive retail landscape, and AI is enabling retailers to deliver highly personalized shopping experiences. Machine learning algorithms analyze customer data, such as browsing history, purchase behavior, and demographic information, to generate personalized product recommendations.

Recommendation engines, such as those used by Amazon and Netflix, leverage collaborative filtering and content-based filtering techniques to suggest products and content that match individual preferences. These systems have been shown to increase customer engagement, reduce bounce rates, and drive sales.

AI is also being used to personalize customer interactions in real-time. For example, chatbots and virtual assistants can use natural language processing to engage customers in personalized conversations, providing product recommendations, answering queries, and assisting with order placements. These AI-driven systems can operate 24/7, providing instant support and improving customer satisfaction.

In addition to enhancing the online shopping experience, AI is being used in physical retail stores to provide personalized services. For example, smart mirrors equipped with AI can analyze customer preferences and suggest clothing items based on their body type and style preferences. These systems enhance the in-store shopping experience and help retailers differentiate themselves from competitors.

Despite the benefits of AI-driven personalization, there are challenges associated with its implementation. Privacy concerns are a significant issue, as the collection and analysis of customer data must be done in compliance with data protection regulations. Additionally, overly personalized recommendations can sometimes feel invasive to customers, potentially leading to a negative shopping experience.

## 5.2 Inventory Management and Demand Forecasting

Effective inventory management is crucial for retail success, and AI is helping retailers optimize their stock levels by predicting demand more accurately. Traditional inventory management systems rely on historical sales data and fixed reorder points, which can result in overstocking or stockouts.

AI-powered demand forecasting models analyze a wide range of data sources, including historical sales, market trends, and external factors such as weather and holidays, to predict future demand for products. This enables retailers to adjust their inventory levels dynamically, reducing carrying costs and minimizing the risk of stockouts.

For example, Walmart has implemented AI-driven demand forecasting to optimize its inventory management, resulting in significant cost savings and improved product availability. Similarly, fashion retailers are using AI to forecast demand for seasonal items, allowing them to adjust their production and ordering strategies accordingly.

In addition to demand forecasting, AI is being used to optimize the replenishment of inventory across retail networks. Machine learning models can analyze sales data, lead times, and supplier performance to determine the optimal reorder quantities and timing. This helps retailers maintain the right inventory levels at each location, reducing stock imbalances and improving supply chain efficiency.

However, the adoption of AI in inventory management requires high-quality data and advanced analytics capabilities. Retailers must invest in data integration and analytics infrastructure to leverage AI effectively. Additionally, the accuracy of AI models can be affected by sudden changes in market conditions, such as economic downturns or shifts in consumer preferences, requiring continuous model updates and validation.

## 5.3 Dynamic Pricing and Revenue Management

Dynamic pricing, also known as price optimization, is a strategy used by retailers to adjust prices in real-time based on factors such as demand, competition, and inventory levels. AI is enabling retailers to implement dynamic pricing strategies more effectively, helping them maximize revenue and improve competitiveness.

Machine learning models can analyze historical sales data, competitor pricing, and customer behavior to identify the optimal price for each product at any given time. This allows retailers to adjust prices dynamically in response to changes in demand or market conditions. For example, online retailers like Amazon use dynamic pricing algorithms to adjust prices for millions of products multiple times a day, ensuring that they remain competitive and maximize sales.

AI-driven dynamic pricing is also being used in physical retail stores. Electronic shelf labels and digital price tags enable retailers to update prices in real-time across multiple locations. This helps retailers respond quickly to changes in market conditions, such as competitor promotions or fluctuations in demand.

In addition to optimizing individual product prices, AI can be used to implement more sophisticated pricing strategies, such as personalized pricing and bundle pricing. Personalized pricing involves offering different prices to different customer segments based on their willingness to pay, while bundle pricing involves offering discounts on complementary products to increase overall sales.

While dynamic pricing offers significant benefits, it also presents challenges. Price changes must be carefully managed to avoid alienating customers or triggering price wars with competitors. Additionally, the use of dynamic pricing requires robust data analytics and pricing optimization capabilities, as well as compliance with pricing regulations.

**Conclusion**

Artificial intelligence is transforming industries by enabling businesses to optimize operations, enhance decision-making, and deliver personalized customer experiences. In healthcare, AI is improving diagnosis and treatment, enabling precision medicine, and transforming healthcare delivery through telemedicine and remote monitoring. In finance, AI is enhancing investment management, fraud detection, and risk management, while in manufacturing, it is enabling predictive maintenance, quality control, and supply chain optimization. In retail, AI is driving personalized recommendations, dynamic pricing, and efficient inventory management.

Despite its potential, the adoption of AI presents several challenges, including data quality and availability, model interpretability, and ethical considerations. Businesses must invest in the necessary infrastructure, talent, and governance frameworks to leverage AI effectively and responsibly. As AI continues to evolve, it will play an increasingly important role in shaping the future of industries, driving innovation, and creating new opportunities for growth and efficiency.